\documentclass[10pt,twocolumn,letterpaper]{article}

\usepackage{soul}
\usepackage{siunitx}
\usepackage{subfigure}
\usepackage{iccv}
\usepackage{times}
\usepackage{epsfig}
\usepackage{graphicx}
\usepackage{amsmath}
\usepackage{nccmath}
\usepackage{amssymb}
\usepackage[fleqn]{mathtools}
\usepackage{bm}
\usepackage{dsfont}
\usepackage{float}
\usepackage{tabularx, booktabs}
\usepackage{calrsfs}
\DeclareMathAlphabet{\pazocal}{OMS}{zplm}{m}{n}

\newcommand{\N}{\mathbb{N}}

\newcommand{\note}[1]{{\color{red}}}
\newcommand{\notea}[1]{{\color{blue}}}
\newcommand{\notej}[1]{{\color{cyan}}}
\newcommand{\notejon}[1]{{\color{magenta}}}
\newcommand\numberthiseq{\addtocounter{equation}{1}\tag{\theequation}}

\usepackage[pagebackref=true,breaklinks=true,letterpaper=true,colorlinks,bookmarks=false]{hyperref}

\iccvfinalcopy %
\def\iccvPaperID{2301} %

\begin{document}

\title{TextTubes for Detecting Curved Text in the Wild}

\author{Jo\"el Seytre\\
\texttt{jsseytre@amazon.com}
\and Jon Wu\\
{\tt jonwu@amazon.com}
\and Alessandro Achille \\
\texttt{aachille@amazon.com}
}

\maketitle

\begin{abstract}
    We present a detector for curved text in natural images. We model scene text instances as tubes around their medial axes and introduce a parametrization-invariant loss function.
    We train a two-stage curved text detector, and evaluate it on the curved text benchmarks CTW-1500 and Total-Text.
    Our approach achieves state-of-the-art results or improves upon them, notably for CTW-1500 by over 8 percentage points in F-score.
\end{abstract}

\section{Introduction}

Detecting and reading text in natural images (also referred to as \textit{scene text} or \textit{text in the wild}) has been a central problem in scene understanding with applications ranging from helping visually impaired people navigate city scenes to product search and retrieval, and instant translation.

Scene text is typically broken down into two successive tasks: (1) text detection attempts to localize characters, words or lines, and (2) text recognition aims to transcribe their content.
Hence, successful text extraction and transcription critically depends on obtaining accurate text localization, which is the main focus of this work.

\begin{figure}[t]
\centering
    \subfigure[]{\includegraphics[width=0.7\linewidth]{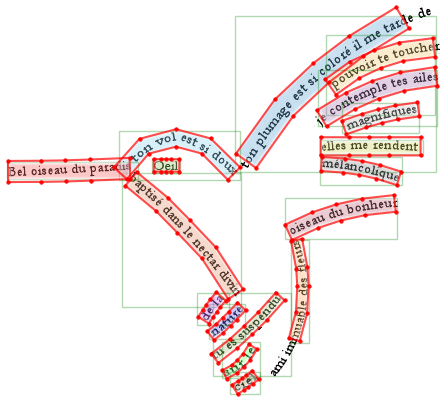}}
    \subfigure[]{\includegraphics[width=.8\linewidth]{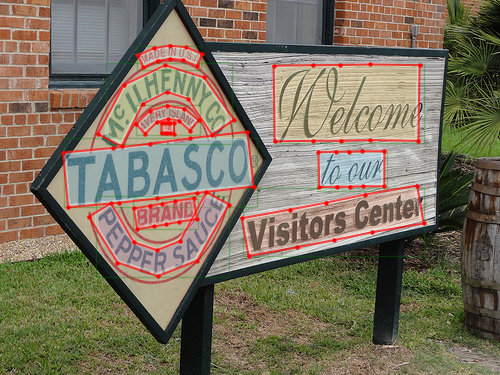}}
    \caption{Inferences using our model and curved text detector TextTubes. Real-life objects often contain nested and curved text, which would be incorrectly retrieved by methods with quadrilateral outputs. (a) Original image  \cite{bird_calligramme} is inspired by Apollinaire's \textit{Calligrammes} \cite{apollinaire2018calligrammes}. (b) is from CTW-1500's test set.}
\label{fig: introductive images}
\end{figure}

Further, text in the wild is affected by large nuisance variability: for example, street signs are deformed by perspective transforms and change of viewpoint, and logos tend to be of arbitrary shapes and fonts. All this makes it difficult to extract \notej{a good standardized reference frame for the} proper detections to feed to the transcription stage of the text processing pipeline.
As Figure~\ref{fig: introductive images} shows, simple quadrilateral frames do not properly capture the text, and in the presence of curved text, the bounding boxes of multiple text instances may overlap. This affects proper retrieval and parsing of the text, since the detection will yield multiple, irregular instances in a frame that should be normalized. 

In this work, we (1) propose a "tube" parametrization of the text reference frame, which can capture most of the nuisance variability through a curved medial axis, which is parametrized by a polygonal chain, alongside a fixed radius for the tube around the medial axis. We then (2) formulate a parametrization-invariant loss function that allows us to train a region-proposal network to detect scene text instances and regress such tubes, while addressing the ambiguous parametrization of the ground-truth polygons. Finally, we (3) achieve state-of-the-art performance on Total-Text and outperform current solutions on CTW-1500 by over eight points in F-score.

\begin{figure*}[t]
\begin{center}
\centering
    \subfigure[Original]{\includegraphics[width=.24\linewidth]{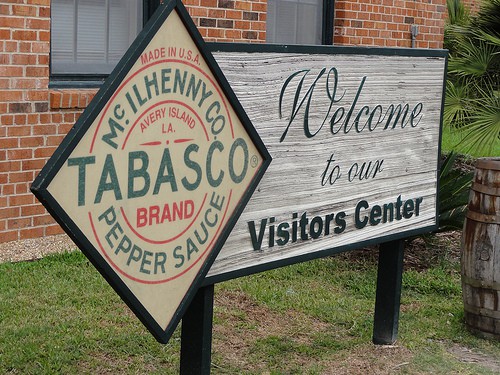}}
    \subfigure[Axis-aligned rectangle]{\includegraphics[width=.24\linewidth]{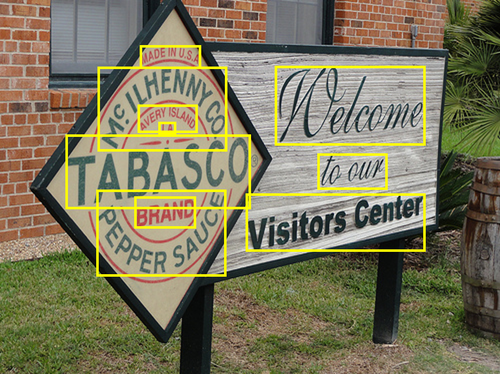}}
    \subfigure[Quadrilateral]{\includegraphics[width=.24\linewidth]{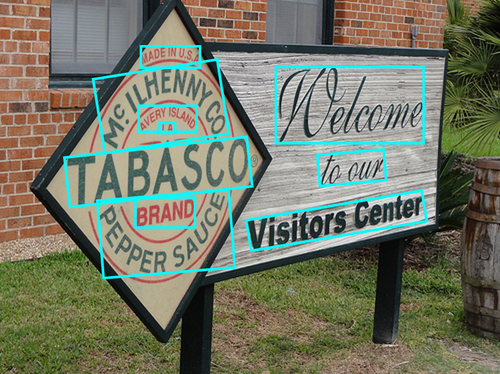}}
    \subfigure[Tube]{\includegraphics[width=.24\linewidth]{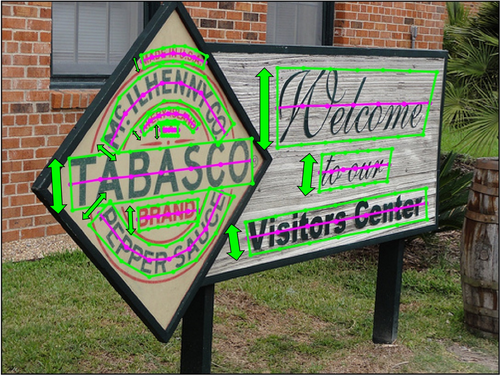}
    \label{fig:tube_subfigure}}
\end{center}
   \caption{Comparing different text representations. (a) shows the original image, (b) and (c) show that rectangles and quadrilaterals generate overlap and often capture as much background noise as text information, while containing several instances in a given box. In (d) the ground truth polygon is in green, and the associated medial axis is in magenta and the arrows represent the radiuses of the tubes.}
\label{fig:snake_representation}
\end{figure*}
\section{Related work}

Early approaches to scene text detection aim to recover single characters in images using Maximally Stable Extremal Regions (MSER) or connected component methods, and then group them into text instances \cite{yin2014robust, yao2012detecting, huang2013text, ye2015text}. False positives are then pruned using AdaBoost classifiers.

More recently, the advent of deep learning has enabled the training of high-accuracy deep networks for text detection, which fall in the following three main categories.

One early approach is to use semantic segmentation to detect text regions and group nearby pixels into full text instances. These methods \cite{FusedText, EAST, zhang2016multi, wu2017self, he2017multi, deng2018pixellink, yue2018boosting, xue2018accurate, liu2018fots} are generally based on Fully Convolutional Networks (FCN) \cite{FCN} or U-Net \cite{U-Net}. The current best-performing method \cite{TextSnake} on the curved text benchmarks CTW-1500 and Total-Text uses a U-Net to segment the text center line, local radius and orientation of the image's text regions, then reconstitutes curved text instances through a striding algorithm.

The second approach is to use a single-stage detector \cite{SSD-reg, liu2017deep, SegLink, tian2016CTPN, tian2017wetext,hu2017wordsup}, such as SSD \cite{SSD} or YOLO \cite{yolo,yolov2}, where the network outputs both the object region and the associated class. Liao \etal \cite{liao2018textboxespp, liao2017textboxes} regress a quadrilateral output from a bounding box proposal, and Bu{\v{s}}ta \etal \cite{buvsta2017deep} achieve text detection and recognition in a single pass.

A third approach is to use two-stage detectors \cite{CTW, jiang2017r2cnn, ma2017arbitrary, liao2018rotation, Jaderberg2014ReadingTextwithCNN, zhu2017deep, yang2018inceptext} such as Faster R-CNN \cite{Faster_R-CNN} or more recently Mask R-CNN \cite{Mask-RCNN}. For example, Sliding Line Point Regression (SLPR) \cite{SLPR} regresses polygonal vertices off a Faster R-CNN output while TextSpotter \cite{TextSpotter} uses Mask R-CNN to derive character-level masks to achieve end-to-end text recognition.
Our approach features a region-proposal stage and falls in this category.

The main challenge of single-stage and two-stage detectors is to successfully separate instances within each region proposals from the Region Proposal Network (RPN), while the key problem for segmentation-based techniques is to accurately separate text instances that might be segmented together during the bottom-up process.

The application of computer vision solutions to a wider set of compelling natural scenes has encouraged the creation of more accurate text datasets: scene text benchmarks have transitioned from axis-aligned rectangles \cite{idcar2003, karatzasidcar2013, veit2016cocotext} to quadrilaterals \cite{karatzasicdar2015, icdar2017-mlt} and most recently to curved text with polygonal ground-truth information \cite{CTW, Total-Text}.

These new curved text benchmarks illustrate the necessity of developing new solutions for curved text. For example, Liu \etal \cite{CTW} report that the popular method EAST \cite{EAST}, which was the first to achieve $80\%$ F-score on the quadrilateral dataset ICDAR 2015, sees a drop in performance from $64\%$ to $35\%$ F-score when comparing performance on the straight and curved text subsets of CTW-1500.

Parametrizing a text instance through its contour vertices is often ambiguous and noisy. On the other hand, an alternative like pixel-level segmentation does not exploit domain knowledge, such as the fact that text is most often distributed along a curve with approximately constant height. To address this, we parametrize text instances as a constant-radius envelope around a text instance's medial axis.

Our proposed method is implemented by generating region proposals as a first step before regressing the region's main text instance's medial axis and associated tube through a parametrization-invariant loss function computed between predicted and ground truth medial axes. This makes it possible for our approach to avoid the noisy and error-prone bottom-up step of grouping pixels into full text instances and ensures correct separation between different instances.

\section{Tube Parametrization}

Our approach represents text instances as follows: we model a tubular region of space through its curved medial axis and a fixed radius along the curve, and design a parametrization-invariant loss which can be used to train a neural network to regress such tubes.

\subsection{Modeling a text instance}
The envelope of a curved text instance may be described as a polygon with $k$ vertices $[x_1, y_1, ..., x_k, y_k] \in \N^{2k}$. This parametrization is the natural extension of quadrilaterals (special case $k=4$) into the more diverse and complex shapes featured in recent datasets \cite{CTW, Total-Text}.
Any given 2-dimensional compact shape can be approximated as a polygon: this increase in precision is just an intermediate step between simple axis-aligned rectangles and full text segmentation, where the number of polygon vertices $k$ is equal to the number of pixels of the shape contour.

Thus, we describe text instances as follows: a curved medial axis and its radius (Fig.~\ref{fig:tube_subfigure}). While tubes are less generic than polygons, they better exploit the tubular shape of text instances.
Notably, this parametrization can be non-unique as the position and number of vertexes can change without affecting the text instance. As described in a subsequent section, our formulation accounts for this variability. 

\paragraph{Medial axis:} The medial axis (sometimes also referred to as the skeleton) of a polygonal text instance is defined as the set of centers of all maximal disks in its polygonal contour \cite{wathen1967models}. %
A disk $\pazocal{D}$ is said to be maximal in an arbitrary shape $\pazocal{A}$ if is is included in $\pazocal{A}$ and any strictly larger disk $\pazocal{D'}$ containing $\pazocal{D}$ would not be included in $\pazocal{A}$.

In any given script, text is usually a concatenation of characters of similar size. 
It follows that a text instance's medial axes should be exclusively non-intersecting polygonal chains, \ie a connected series of line segments, and we compute them as such. We also extend the two end segments of the polygonal chains until they join the polygon, as shown in Figure~\ref{fig:snake_representation}. The ground-truth medial axis of a polygonal text instance with $2k$ vertices is thus described by $k$ points that form the $k-1$ segments of the medial axis when ordered properly. In the rest of this paper we will refer to these $k$ vertices along the medial axis as the \textit{medial points} of the text instance. This is done in order to differentiate them from generic keypoints (see also Section \ref{section: detectron keypoints}).

\paragraph{Radius:} as illustrated on Fig~\ref{fig:tube_subfigure} we define the radius of a given text instance as the average disk radius over the set of all maximal disks in the polygonal text region. In our model, we make the approximation that a given tube's radius is constant.
As shown in Fig \ref{fig: datasets} this is a reasonable assumption as roughly 4 out of 5 instances from both CTW-1500 and Total-Text have less than a $20\%$ variation in radius along their medial axis, which results in a Intersection-over-Union of more than $0.9$ between a fixed-radius tube model and varying-radius tube.

\subsection{Training loss function}

\paragraph{Multi-task loss:}
Similarly to the original approach of He \etal \cite{Mask-RCNN} we train the network using the multi-task loss
\begin{equation}
\ell = \ell_{\text{cls}} + \ell_{\text{bbox}} + \ell_{\text{tube}},
\label{eq: multi-task loss}
\end{equation}
where $\ell_{\text{cls}}$ is the cross-entropy loss between predicted and ground truth class (\textit{text} or \textit{no text}) which is determined based on the Intersection-over-Union (IoU) to the ground truth bounding boxes in the image, and $\ell_{\text{bbox}}$ is an $L_1$ loss between the coordinates of the predicted bounding box and those of the rectangular, axis-aligned ground truth box encompassing the polygonal text instance.
\begin{figure}[t]
\begin{center}
\includegraphics[width=\linewidth]{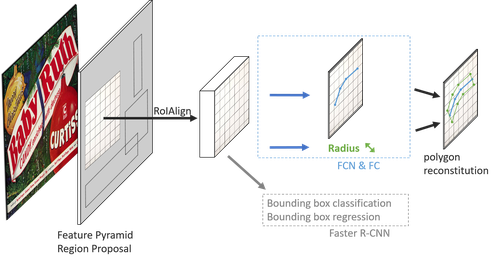}
\end{center}
   \caption{TextTubes's architecture is built on top of the Mask R-CNN model, where we replace the upsampling of the mask head with the direct regression of the medial points and the text radius through 2 fully connected layers of 256 neurons.}
\label{fig: architecture}
\end{figure}

\paragraph{Tube loss:} The purpose of our proposed loss function $\ell_{\text{tube}}$ is to align predicted and ground truth tubes' medial axes. The network outputs points that can be mapped to a curve parametrization, which is at the heart of the model. 
We avoid naively introducing a regression loss on each individual medial keypoint, since the precise choice of medial points is ambiguous and not unique, and the network may be unfairly penalized for finding equivalent parametrizations. This leads to unstable training and worse overall performance (see Section \ref{section: detectron keypoints}).
Rather, we design a parametrization invariant loss, which accounts for both proximity of the regressed medial axis to the ground-truth, and proximity of their gradients (similar to a Sobolev norm).

Let $\hat{\bm{S}} = (x_1, y_1, \ldots, x_{\hat{n}}, y_{\hat{n}})$ be the output of the network and $\bm{S} = (x_1, y_1, \ldots, x_{\hat{n}}, y_{n})$ be the ground truth medial axis, where $\hat{n}$ and $n$ are the number of medial points of the output and of the ground truth medial axis respectively. The loss $\ell_{\text{tube}}$ is defined as
\begin{equation}
\ell_{\text{tube}} = \ell_{\text{radius}} + \ell_{\text{axis}} + \ell_\text{{endpoints}} + \ell_\text{{spread}}.
\label{eq: snake-loss}
\end{equation}

$\ell_{\text{radius}}$ is a standard $L_1$ loss function between each instance's average ground truth radius and predicted radius.

$\ell_\text{{endpoints}}$ is also an $L_1$ loss for the initial and final medial points. While intermediate medial points are not unique due to reparametrization invariance, the endpoints of the tube are \textit{not} ambiguous, and should be regressed accurately.

The remainder of the medial points only serve as support for the underlying tube and their precise location does not affect the parametrization invariant loss $\ell_{\text{axis}}$. To ensure that medial points capture the overall instance medial axis and avoid having medial points collapse together, we define a repulsive elastic loss between successive medial points.
\[
\ell_{\text{spread}} = 
\begin{cases}
d_\text{threshold} - d_\text{min} \text{\hspace{0.5cm} if $d_\text{min} < d_\text{threshold}$}\\
0                                   \text{\hspace{2.5cm} otherwise,}
\end{cases}
\numberthiseq
\]
where $d_\text{min}$ is the length of the smallest segment of the medial axis $\hat{\bm{S}}$ \textit{i.e} the distance between its two closest medial points and $d_\text{threshold} = \frac{l}{2(\hat{n} - 1)}$ where $l$ denotes the total length of the ground truth medial axis.

\paragraph{Line loss}
We formulate $\ell_{\text{axis}}$ as follows:
\begin{align*}
\ell_{\text{axis}}(\hat{\bm{S}}, \bm{S})  \vcentcolon= 1 &- \hspace{3mm} \alpha \  \mathcal{S}_{\text{abs}}(\hat{\gamma}, \gamma) \\
&- (1 - \alpha) \mathcal{S}_{\text{tan}}(\hat{\gamma}, \gamma) \numberthiseq
 \label{eq: line-loss}
\end{align*}
where $\hat{\gamma},\gamma: [0,1] \to \mathbb{R}^2$ are arc-length parametrizations of predicted and ground truth polygonal chains $\hat{\bm{S}}$ and $\bm{S}$, and 
\begin{align*}
\begin{cases}
\mathcal{S}_{\text{abs}}(\hat{\gamma},\gamma) &\vcentcolon= \int_0^1 \exp(- \frac{d(\hat{\gamma}(t), \gamma)^2}{2 \sigma^2})  dt\\ 
\mathcal{S}_\text{tan}(\hat{\gamma},\gamma) &\vcentcolon= \int_0^1 \exp(- \frac{\sin^2|\hat{\theta}(t) - \theta(t)|}{2 \sigma'^2}) dt.
\end{cases}\numberthiseq \label{eq: similarities}
\end{align*}
where $\mathcal{S}_{\text{abs}}$ measures the overall proximity between the two medial axes, and $\mathcal{S}_{\text{tangents}}$ measures the  similarity between directions. Here, $d(p, \gamma)$ measures the distance from the point $p$ to the curve $\gamma$, and $\hat{\theta}$ and $\theta$ are the angles of the prediction and ground truth tangents respectively at point $\hat{\gamma}(t)$ and at its closest point $p \in \gamma$ such that $d(\hat{\gamma}(t), p) = d(\hat{\gamma}(t), \gamma)$.
In practice, we approximate this loss function numerically by sampling points uniformly across $\hat{\bm{S}}$ and $\bm{S}$.

The loss function introduces several hyper-parameters. We choose to sample 100 points uniformly along the medial axis to compute distances and tangent differences. The parameter $\alpha \in [0,1]$ interpolates between giving full weight to the distance between curves ($\alpha=0$), or to the distance between their derivatives ($\alpha=1$). We found $\alpha=0.5$ to give the best performance, and $\sigma$ and $\sigma'$ are set to the standard variations of the Gaussian kernels. The latter two are chosen based on the scale of the inputs to the Gaussian kernels $\mathcal{S}_{\text{abs}}$ and $\mathcal{S}_{\text{tan}}$. In addition to these hyper-parameters, the loss terms in Eq.~\ref{eq: multi-task loss} and \ref{eq: snake-loss} can be assigned individual weights: we cross-validate them to be equal.
In order to have a fixed size output from our model we fix the number of medial points to a hyper-parameter $\hat{n}$.
We found that TextTubes's performance is independant of the number of medial points regressed, when using at least $\hat{n}=4$ medial points, as the medial axis can only consist of $2$ segments or less if there are 3 or less medial points. More specifically, the difference in performance between 4, 5, 7, 10 and 15 medial points is less than 0.5 percentage points in F-score. In the rest of this study we take $\hat{n}=5$.

\begin{figure}[b]
\centering
  \subfigure[CTW-1500]{\includegraphics[width=0.75\linewidth]{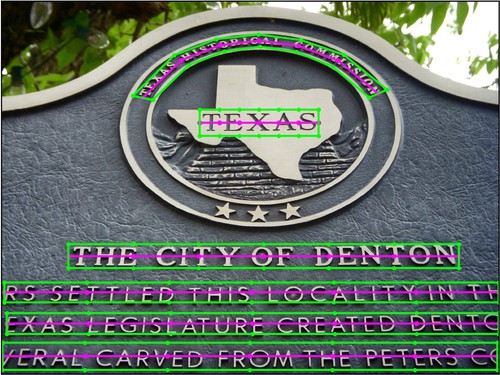}}
  \subfigure[Total-Text]{\includegraphics[width=.9\linewidth]{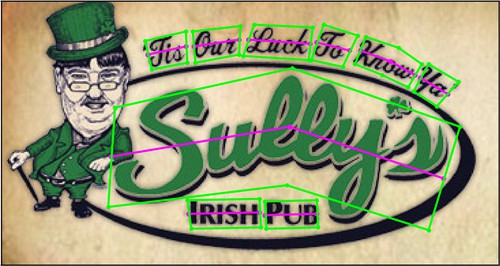}}
\caption{Comparison of ground truth bounding polygons: these images show the difference between CTW-1500's line-level information and Total-Text's word-level bounding polygons.}
\label{fig:datasets_gt_comparison}
\end{figure}
\begin{figure*}[t]
    \centering
    \begin{tabular}[t]{cc}
    \resizebox{1.0\columnwidth}{!}{
    
    \begin{tabular}[t]{c}
        \footnotesize
        \begin{tabular}{clcc}
            \toprule
            & dataset & CTW-1500 & Total-Text \\
            &text information                   & lines             & words     \\
            \hline
            &images                             & 1,000             & 1,255     \\
            train&straight instances            & 4,655             & 6,667     \\
            &curved instances                   & 3,037             & 3,912     \\
            &total instances                    & 7,692             & 10,579    \\   
            \hline
            &images                             & 500               & 300       \\
            test&straight instances             & 1,535             & 1,592     \\
            & curved instances                  & 1,533             & 954       \\
            &total instances                    & 3,068             & 2,546     \\
            \hline
            &images                             & 1,500             & 1,555     \\
            overall&straight instances          & 6,190             & 8,259     \\
            &curved instances                   & 4,570             & 4,866     \\
            &total instances                    & 10,760            & 13,125    \\
            \bottomrule
        \end{tabular}
        \\
        \subfigure[Details of the image and instance count.]{\rule{5cm}{0cm}\label{table: datasets}}
    \end{tabular}
    }
    
        &
    
    \resizebox{1.0\columnwidth}{!}{
    \begin{tabular}{c}%
        \smallskip
        \subfigure[CTW-1500: curvature]{\includegraphics[width=0.29\linewidth]{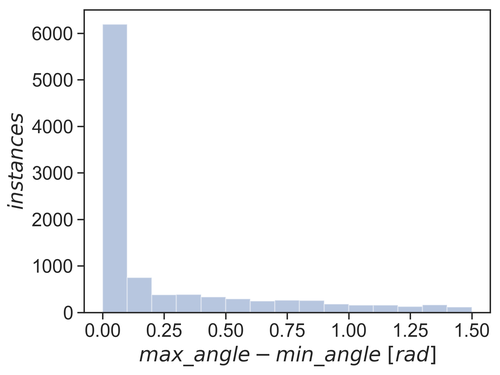}}
        \subfigure[CTW-1500: radius variation]{\includegraphics[width=0.29\linewidth]{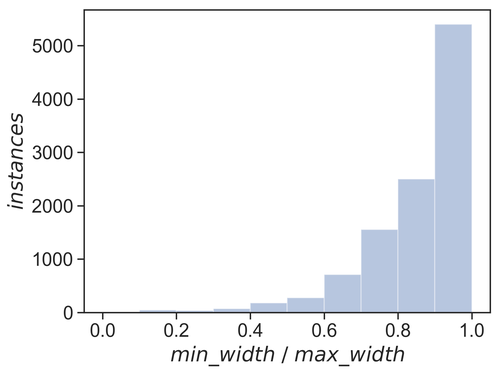}}\\
        \subfigure[Total-Text: curvature]{\includegraphics[width=0.29\linewidth]{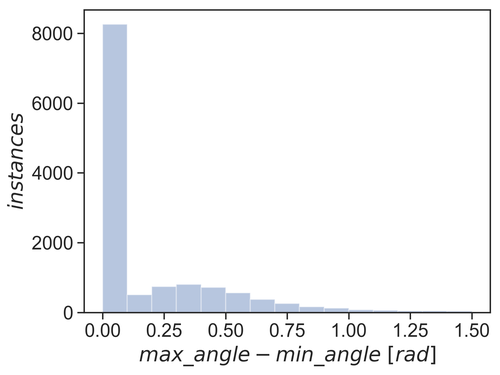}}
        \subfigure[Total-Text: radius variation]{\includegraphics[width=0.29\linewidth]{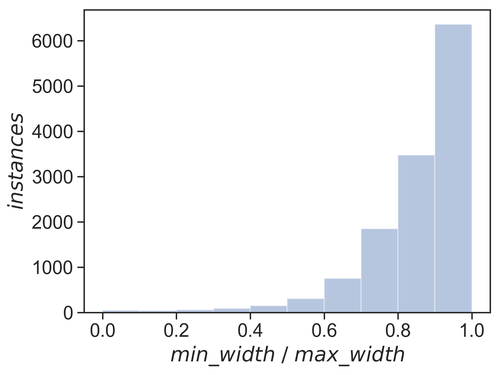}}
    \end{tabular}
    }
    \end{tabular}
    
    \caption{Dataset information. (a) breaks down the number of images and instance. (b) and (d) are histograms on the maximal angle difference between medial axis segments whereas (c) and (e) describe the relative radius variation along a text instance. Notable difference: CTW-1500 has instances that curve more than Total-Text. This is due to the text instance length difference (see Section \ref{section: curved text datasets} and Fig \ref{fig:datasets_gt_comparison}).}
    \label{fig: datasets}
\end{figure*}
\subsection{Computing a tube using a deep network}

As described in Figure~\ref{fig: architecture}, we build on top of Mask R-CNN's \cite{Mask-RCNN}: the input image goes through a Region-Proposal Network (RPN) from which Regions of Interest (RoI) features are computed using a Feature Pyramid Network (FPN) \cite{FPN}. Those RoI features go through the Faster R-CNN detector \cite{Faster_R-CNN} to determine the class and bounding box of the instance. In parallel, the RoI features are also fed to our newly introduced tube branch, which consists of a Fully Convolutional Network (FCN) \cite{FCN} on top of which the coordinates of the $\hat{n}$ medial points as well as the radius of the text instance are regressed.
Implementation details are given in Section \ref{section: training}.
All the points along the medial axes are linear interpolations of the vertexes output from the network, and therefore can be sampled while retaining differentiability w.r.t. the network output.

\section{Datasets and evaluation}
\subsection{Curved text datasets} \label{section: curved text datasets}

\paragraph{CTW-1500} is a dataset composed of 1,500 images that were collected from natural scenes, websites and image libraries like OpenImages \cite{openimages}. It contains over 10,000 text instances with at least one curved instance per image.
The ground truth polygons are annotated in the form of 14-vertex polygons with line-level information.

\paragraph{Total-Text} is a dataset composed of 1,255 train images and 300 test images that can contain curved text instances. 
The ground truth information is in the form of word-level polygons with a varying number of vertices, from 4 to 18.

\paragraph{Curved and straight subsets}
To address the importance of our curved text contribution, we distinguish whether individual text instances are curved or straight and form a curve and a straight subset of each dataset.  
Total-Text distinguishes curved instances in its annotations without specifying how, whereas CTW-1500 does not provide this information. Thus we apply our own criteria to both datasets.

We determine whether or not an instance is curved based on its medial axis: it is curved if any two distinct segments vary in angle by more than $0.1$ radian ($\approx \ang{6}$). This yields roughly half of curved instances for CTW-1500 and one third for Total-Text, as seen in Fig \ref{table: datasets}.
For more details we refer the reader to the original datasets paper \cite{CTW, Total-Text}.

\subsection{Evaluation protocol}
We base our evaluations on the polygonal PASCAL VOC protocol \cite{everingham2010pascal} at Intersection-over-Union (IoU) $= 0.5$ threshold, as made publicly available \footnote{\url{https://github.com/Yuliang-Liu/Curve-Text-Detector/tree/master/tools/ctw1500_evaluation}} by Lyu \etal \cite{CTW}.

After ranking polygonal predictions based on their confidence score, predictions are true positives if their IoU with a ground-truth polygon is greater than $0.5$. Any ground-truth instance can only be matched once.
The Precision-Recall (PR) curves associated to TextTubes are shown on Figure~\ref{fig: p-r curve}.
From such PR curves we can extract the precision and recall corresponding to the maximum F-score.

\begin{table*}[t]
\footnotesize
\begin{center}
\resizebox{1.8\columnwidth}{!}{
\begin{tabular}{cccccccc}
\toprule
                & Additional Text Images & \multicolumn{3}{c}{CTW-1500}                             & \multicolumn{3}{c}{Total-Text} \\
Method          & Used For Training & P (\%)              & R (\%)      & F (\%)       & P (\%)              & R (\%)      & F (\%)      \\
\hline
Tian \etal (2016) \cite{tian2016CTPN} &  & $60.4^\dagger$ & $53.8^\dagger$ & $56.9^\dagger$ & - & - & - \\
Shi \etal (2017) \cite{SegLink}&  & $42.3^\dagger$ & $40.0^\dagger$ & $40.8^\dagger$ & $30.3^\ddagger$ & $23.8^\ddagger$ & $26.7^\ddagger$\\
Zhou \etal (2017) \cite{EAST}& & $78.7^\dagger$& $49.1^\dagger$& $60.4^\dagger$& $50.0^\ddagger$& $36.2^\ddagger$& $42.0^\ddagger$\\
\hline
Liu \etal (2017) \cite{CTW} &                       & 77.4      & 69.8      & 73.4              & -     & -     & -             \\
Ch'ng \etal (2017) \cite{Total-Text} & $\approx 15$ K (COCO) & -         & -         & -                 & 33    & 40    & 36            \\
\hline
Zhu \etal (2018) \cite{SLPR}        &               & 80.1      & 70.1      & 74.8              & -     & -     & -             \\
Lyu \etal (2018) \cite{TextSpotter} & $\approx 800$ K (synth. + ICDAR13/15)  & -         & -         & -                 & 69.0  & 55.0  & 61.3          \\
Long \etal (2018) \cite{TextSnake}  & $\approx 800$ K (synth.)  & 67.9      & 85.3      & 75.6              & 82.7  & 74.5  & 78.4          \\
\hline                 
$\text{Baseline}^{\star}$ &                         & 78.9      & 73.7      & 76.2              & -     &  -    & -             \\
TextTubes (no pre-training) &                                     & 83.54     & 79.00     & 81.21             & 77.56 &  73.00& 75.21         \\
TextTubes$^{(*)}$& $\approx 1$ K                                  & 87.65     & 80.00     & \textbf{83.65}    & 84.25 & 73.50 & \textbf{78.51}\\
\bottomrule
\end{tabular}
}
\end{center}
\caption[caption]{Comparing Precision, Recall and F-score on CTW-1500 and Total-Text, evaluated at Intersection-over-Union (IoU) $=0.5$.\\\hspace{5cm}
\textsuperscript{$\dagger$} Results for quadrilateral methods on CTW-1500 are taken from Liu \etal \cite{CTW} and trained on CTW-1500's circumscribed rectangles.\\\hspace{5cm}
\textsuperscript{$\ddagger$} Results for these methods are taken from Long \etal \cite{TextSnake} and are not fine-tuned on Total-Text.\\\hspace{5cm}
\textsuperscript{$\star$}This keypoints baseline is not applicable to Total-Text because of its varying number of vertices per ground truth polygon (see Sec. \ref{section: detectron keypoints}).\\\hspace{5cm}
\textsuperscript{$(*)$} We initialize the network with weights learned from training on CTW-1500 for Total-Text and \textit{vice versa}.
}
\label{table: results}
\end{table*}
\subsection{Dataset difference: instance-level information}
Although Total-Text and CTW-1500 are both relevant to the study of curved text, the two datasets offer different levels of granularity in their ground truth information.
As displayed in Figure~\ref{fig:datasets_gt_comparison}, CTW-1500 groups words together if they are aligned, whereas Total-Text uses a different polygon for each individual word.

\section{Experiments}
In this section we evaluate and compare our tube parametrization and our trained text detector to the established benchmarks of the scene text detection literature.

\subsection{Training}\label{section: training}

We initialize our network with a ResNet-50 backbone pre-trained on ImageNet.
We train for a total of $15,000$ steps with minibatches of 2 images from which we extract 512 RoI each. We use Stochastic Gradient Descent (SGD): during a warm-up period of $1,000$ steps we ramp up the learning rate from one third to its full  value of $0.005$, which is divided by 10 at steps $8,000$ and $12,000$.
We use $0.0001$ for weight decay and $0.9$ for momentum.
Our region proposals' aspect ratios are sampled in $\{0.2, 0.5, 1, 2, 4\}$ and we use levels 2 to 6 of the FPN pyramid.

During training we randomly resize the image to values in the [640, 800] pixel range.
For inference we resize the image so that its shortest side is 800 if we are using mono-scale. For multi-scale inference we resize images from CTW-1500 to $(400, 600, 1000)$ and Total-Text to $(400, 800, 1200)$.
To improve performance on rotated and vertical text, we also augment the data by applying a random 90 degrees rotation with probability $=0.4$.
\vspace{5mm}
\begin{figure}[b]
\centering
\subfigure{\includegraphics[width=0.495\linewidth]{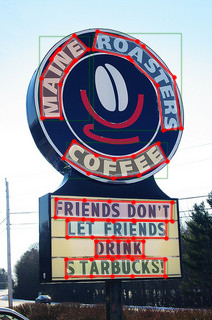}}\hspace{0cm}
\subfigure{\includegraphics[width=0.481\linewidth]{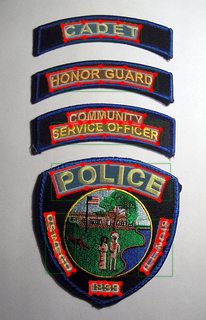}}
\caption{Example of tube inferences: we accurately capture both the straight and curved text regions with various orientations.}
\end{figure}
\subsection{Post-Processing}
\paragraph{Non-Maximal Suppression (NMS)}
As is common in object detection frameworks, we use soft-NMS \cite{Soft-NMS} at the IoU $= 0.5$ threshold on the rectangular outputs of the Faster R-CNN module, in order to suppress similar region-proposals and reduce the number of overlapping predictions that result in false positives. Soft-NMS results in a better performance than hard NMS as those boxes are the base for the regression of separate tubes that might share overlapping rectangular bounding boxes, \eg in the case of real life nested text instances such as those of Fig. \ref{fig: introductive images}.

We want to avoid tubes that would overlap on close-by text instances: on top of bounding box soft-NMS, we use the traditional hard NMS between output tubes by computing their polygonal intersections.

\subsection{Results}

We achieve state-of-the-art results on CTW-1500 and Total-Text as shown in Table \ref{table: results}.
While we achieve state-of-the-art for Total-Text, our performance is significantly better on CTW-1500 as the closest method achieves $75.6\%$ F-score whereas TextTubes's best results reach $83.65\%$.
We address our understanding of the gap difference in Section \ref{section: discussion}.
Average inference time on a V100 GPU is $358$ ms with multi-scale and $153$ ms without.

\notej{Comments on Total-Text?}
\begin{figure}[t]
\centering
\subfigure[CTW-1500]{\includegraphics[width=0.99\linewidth]{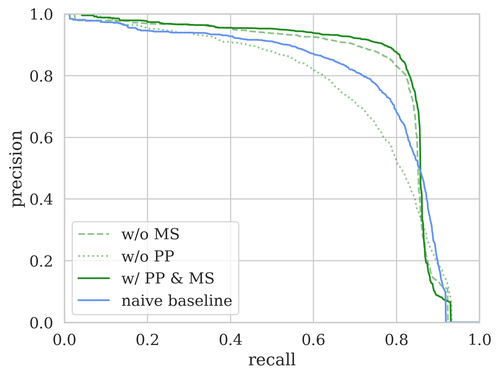}}
\subfigure[Total-Text]{\includegraphics[width=0.99\linewidth]{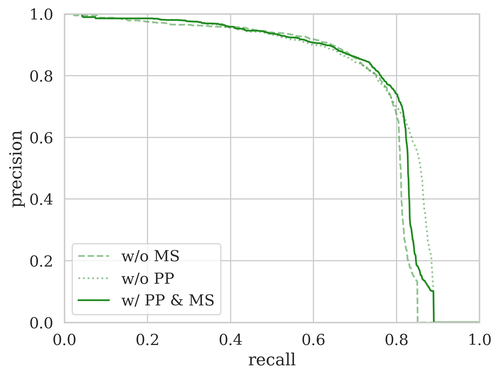}}
\caption{Precision-Recall Curves, comparing the best results with the removal of polygonal post-processing (PP) and multi-scale (MS) and Section~\ref{section: detectron keypoints}'s baseline keypoints approach.}
\label{fig: p-r curve}
\end{figure}
\subsection{Ablation study}
As we can see in Table \ref{table: ablation} and in Figure~\ref{fig: p-r curve} we measure the difference in performance of TextTubes while removing the tube-specific polygonal post-processing and multi-scale, as well as TextTubes's results on the curve and straight subsets (as defined in Sec. \ref{section: curved text datasets}) of both datasets.

The Precision-Recall (PR) curves suggest that the impact of multi-scale is not very significant. On the other hand, post-processing has a large impact, specifically on CTW-1500.
The impact of post-processing on CTW-1500 performance may be attributed to (1) the fact that CTW-1500's line-level instances result in many more overlapping tube outputs than Total-Text's word level predictions, and that accordingly (2) line-level polygonal tubes differ more from the Faster R-CNN output boxes than word-level tubes, which in turn causes polygonal NMS to impact CTW-1500 more than Total-Text.

Furthermore, the post-processing helps the network retain similar region-proposals that result in non-overlapping tube, \eg in the case of nested instances (Fig~\ref{fig: introductive images}), while discarding similar region-proposals that are processed into tubes for the same text instances.

\subsection{Measuring performance on straight text}
It is of paramount importance to assess whether our model performs well on the specific task of curved text detection, while also detecting straight text accurately. This is why we assess our method on the curved and straight subsets of CTW-1500 and Total-Text in Table \ref{table: ablation}.
The performance on the curved and straight subsets cannot include neither precision nor F-score, as the other subset is discounted from the ground truth. Our ablation study indicates that our model captures $90.5\%$ of the CTW-1500's curved instances within an IoU $=0.5$ margin, while reaching up to $95.6\%$ of straight instances. The fact that TextTubes recovers seven points more of Total-Text's curved instances than straight is due to the dataset containing almost twice as many straight instances, including many very short words composed of few characters that result in false negatives.

Achieving similar results on the curved and straight subsets illustrates the robustness of our method.

\subsection{Baseline: learning polygon vertices independently} \label{section: detectron keypoints}

\begin{figure*}[t]
\centering

  \subfigure{\includegraphics[width=0.253\linewidth]{images/CTW_1051_GT.jpg}}
  \subfigure{\includegraphics[width=0.253\linewidth]{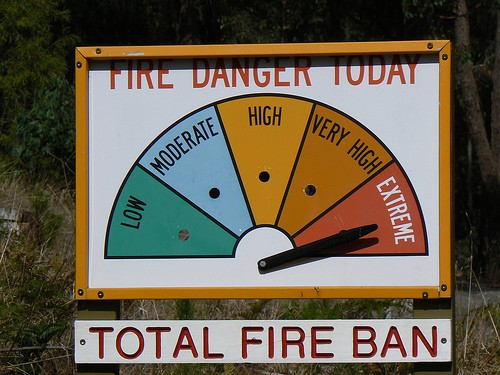}}
  \subfigure{\includegraphics[width=0.1912\linewidth]{images/CTW_1171_GT.jpg}}
  \subfigure{\includegraphics[width=0.278\linewidth]{images/CTW_1104_GT.jpg}}\\
  \subfigure{\includegraphics[width=0.253\linewidth]{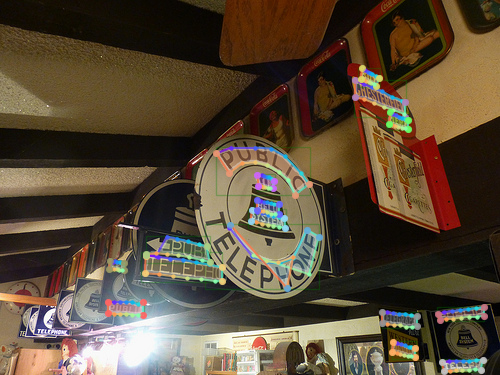}}
  \subfigure{\includegraphics[width=0.253\linewidth]{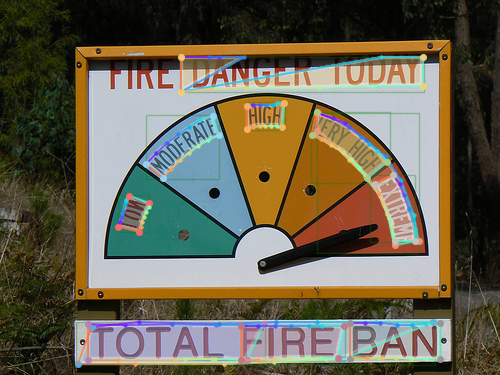}}
  \subfigure{\includegraphics[width=0.1912\linewidth]{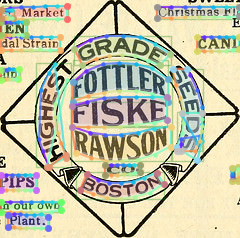}}
  \subfigure{\includegraphics[width=0.278\linewidth]{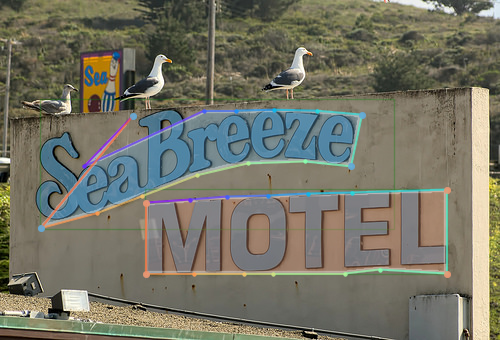}}\\
  \subfigure{\includegraphics[width=0.253\linewidth]{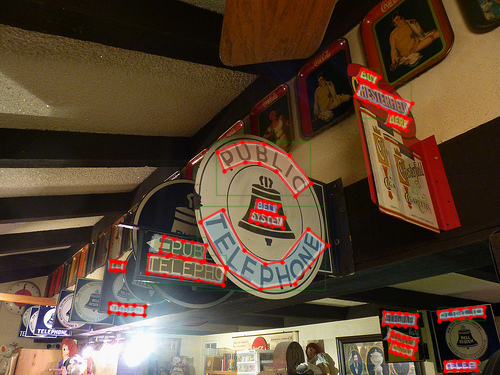}}
  \subfigure{\includegraphics[width=0.253\linewidth]{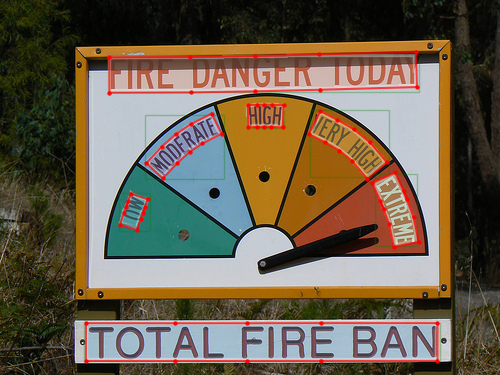}}
  \subfigure{\includegraphics[width=0.1912\linewidth]{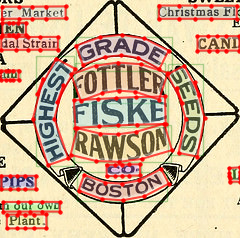}}
  \subfigure{\includegraphics[width=0.278\linewidth]{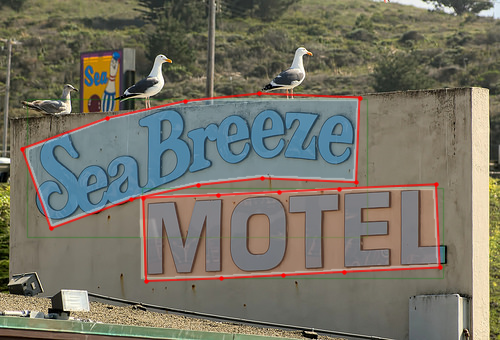}} \\
   \caption{Inferences. The first row contains the original images, the second row shows the na\"ive baseline results (see Sec. \ref{section: detectron keypoints}) and the third is TextTubes's outputs. Treating each vertex as an individual keypoint often causes the detector to miss because of one keypoint being located on the wrong part of an accurate region proposal, \eg for the rightmost picture the top-left keypoint is wrongly located on the "B". We color each keypoint differently in the second row to make this more visual.}
\label{fig: detectron keypoints}
\end{figure*}

A relevant baseline to consider is to naively use the Mask R-CNN architecture to estimate the vertices of the ground-truth polygon as individual keypoints, rather than outputting medial points that follow our tube parametrization through TextTubes. Our comparison baseline has only limited applicability, since it assumes a fixed number of points on the ground-truth, which, while true for CTW-1500, does not hold for other datasets (\eg Total-Text).
In Table \ref{table: results}, we show that TextTubes outperforms this approach by $7.45$ points in F-score.
This may be due to TextTubes's loss function being invariant to reparametrization, while the na\"ive baseline loss, being defined independently on each point, does not capture global properties of the text instance.
Fig. \ref{fig: detectron keypoints} shows for several images that the na\"ive approach often sees one of the 14 vertices as wrongly predicted, which in turn causes the entire prediction to be missed, \eg by having intersecting sides. The increase in performance when comparing this na\"ive baseline to TextTubes results showcases the advantage of using our introduced tube loss.

\begin{table}
\Huge
\begin{center}
\resizebox{1\columnwidth}{!}{
\begin{tabular}{ccccccccc}
\toprule
Ablation study              & \multicolumn{4}{c}{CTW-1500}          & \multicolumn{4}{c}{Total-Text}     \\
                            & $\text{R}_{\text{max}}$ (\%)& P (\%)& R (\%)& F (\%)    & $\text{R}_{\text{max}}$ (\%)    & P (\%)    & R (\%)    & F (\%)    \\
\hline
straight subset             & 95.60     & -     & -     & -         & 86.24         & -         & -         & -         \\
curved subset               & 90.50     & -     & -     & -         & 93.71         & -         & -         & -         \\
\hline
no post-process.            & 93.20     & 72.5  & 69.1  & 70.8      & 88.90         & 81.13     & 74.30     & 77.57     \\
no multi-scale              & 92.30     & 84.8  & 79.0  & 81.8      & 85.00         & 81.36     & 74.20     & 77.62     \\
Best results                & 93.00     & 87.65 & 80.00 & 83.65     & 88.90         & 84.25     & 73.50     & 78.51     \\
\bottomrule
\end{tabular}
}
\end{center}
\caption{Ablation study of TextTubes on CTW-1500 and Total-Text. No post-processing indicates the removal of polygonal NMS. We do not feature the precision (and F-score) for the straight and curved subset because of false positives from excluded subsets. $\text{R}_{\text{max}}$ describes what percentage of the datasets' test instances we manage to capture.}
\label{table: ablation}
\end{table}

\section{Discussion} \label{section: discussion}
Modeling an instance's medial axis and average radius instead of directly inferring the associated polygon achieves two main goals: (1) it is a robust way to compute a loss function on the text instance's shape (it wouldn't be obvious how to compute a polygon-based loss without computing the loss on a vertex-to-vertex basis); (2) it captures information about the instance overall and does not overfit to a given dataset's choice of representation, which would cause the features to be locally restricted to specific keypoints-to-keypoints mapping along the text instances.

Further, we would like to highlight how our model performs on text instances that are individual words vs. lines of words.
On datasets that consist of individual words, such as Total-Text, our model is able to achieve state-of-the-art performance. 
On datasets that have line-level annotations, such as CTW-1500, our model is able to better capture textual information along an instance's separate words. 
This improvement can be seen in our results on CTW-1500, where we have a significant improvement over state-of-the-art.

We can foresee further quality and precision improvements to our method by improving the underlying region proposal network (based on Faster R-CNN). Though it is typically sufficient to use axis-aligned boxes for scene text, even better performance could possibility be achieved with the addition of our tube-loss by including rotated bounding box proposals (e.g. through rotated anchor/default-boxes) as shown in the work of Ma \etal \cite{ma2017arbitrary}.

\section{Conclusion}

We propose a tube representation for scene text instances of various shapes and train a TextTubes curved text detector to learn it. This two-stage network regresses the medial axis and the radius of the text instances for each Region-of-Interest of the region-proposal module, from which an accurate tube is computed.
Our method sees improvements on state-of-the-art results on the established curved text benchmarks CTW-1500 and Total-Text.

While this tube representation is particularly relevant for text instances, it could be adapted to other tasks where the medial axis is complex yet relevant, such as pose estimation. 

\clearpage
{\small
\bibliographystyle{ieee}
\bibliography{egbib}
}

\end{document}